\documentclass[twoside,11pt]{article}

\usepackage{dnd}
\usepackage{times}
\usepackage{multirow,amsmath, bm} 
\usepackage{algorithm2e} 

\jmlrheading{?(?)}{2015}{??-??}{?/??}{?/??}{?/??}{Kai Sun, Qizhe Xie,  and Kai Yu}{10.5087/dad.2015.???}

\ShortHeadings{Recurrent Polynomial Network for Dialogue State Tracking}{{Sun}, Xie and Yu}
\firstpageno{1}

\begin{document}

\title{Recurrent Polynomial Network for Dialogue State Tracking\thanks{This work was supported by the Program for Professor of Special Appointment (Eastern Scholar) at Shanghai Institutions of Higher Learning and the China NSFC project No. 61222208.}}

\author{\name Kai Sun \email accreator@sjtu.edu.cn \\
       \addr Key Laboratory of Shanghai Education Commission for Intelligent Interaction and Cognitive Engineering\\
       SpeechLab, Department of Computer Science and Engineering\\
       Shanghai Jiao Tong University, Shanghai, 200240, China\\
       \AND
       \name Qizhe Xie \email cheezer@sjtu.edu.cn \\
       \addr Key Laboratory of Shanghai Education Commission for Intelligent Interaction and Cognitive Engineering\\
       SpeechLab, Department of Computer Science and Engineering\\
       Shanghai Jiao Tong University, Shanghai, 200240, China\\
       \AND
       \name Kai Yu\thanks{Corresponding author} \email kai.yu@sjtu.edu.cn\\
       \addr Key Laboratory of Shanghai Education Commission for Intelligent Interaction and Cognitive Engineering\\
       SpeechLab, Department of Computer Science and Engineering\\
       Shanghai Jiao Tong University, Shanghai, 200240, China\\
       }

\editor{???}

\maketitle

\begin{abstract}%
  Dialogue state tracking (DST) is a process to estimate the distribution of the dialogue states as a dialogue progresses.  Recent studies on constrained Markov Bayesian polynomial (CMBP) framework take the first step towards bridging the gap between rule-based and statistical approaches for DST.  In this paper, the gap is further bridged by a novel hybrid framework -- \emph{recurrent polynomial network }(RPN).  RPN's unique structure enables the framework to have all the advantages of CMBP including efficiency, portability and interpretability.  Additionally, RPN achieves more properties of statistical approaches than CMBP.  RPN was evaluated on the data corpora of the second and the third Dialog State Tracking Challenge (DSTC-2/3).  Experiments showed that RPN can significantly outperform both traditional rule-based approaches and statistical approaches with similar feature set.  Compared with the state-of-the-art statistical DST approaches with a lot richer features, RPN is also competitive.
\end{abstract}


\section{Introduction}
\label{sec:introduction}
\begin{figure}[htbp!]
\centering
\includegraphics[width=14cm]{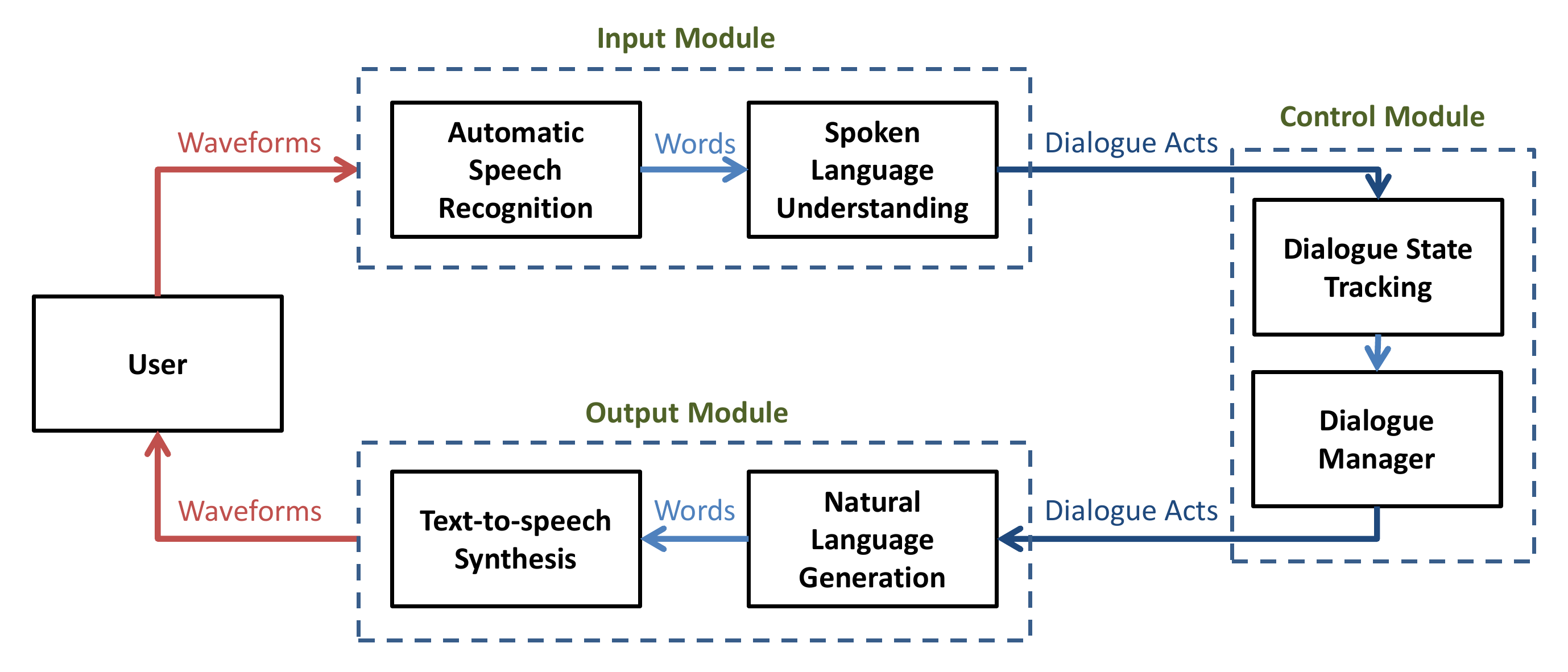}
\caption{Diagram of a spoken dialogue system (SDS)}
\label{fig:intro:sds}
\end{figure}

A task-oriented spoken dialogue system (SDS) is a system that can interact with a user to accomplish a predefined task through speech. It usually has three modules: input, output and control, shown in Figure \ref{fig:intro:sds}.  The input module consists of automatic speech recognition (ASR) and spoken language understanding (SLU), with which the user speech is converted into text and semantics-level user dialogue acts are extracted.  Once the user dialogue acts are received, the control module, also called dialogue management accomplishes two missions.  One mission is called dialogue state tracking (DST), which is a process to estimate the distribution of the dialogue states, an encoding of the machine's understanding about the conversion as a dialogue progresses.  Another mission is to choose semantics-level machine dialogue acts to direct the dialogue given the information of the dialogue state, referred to as dialogue decision making.  The output module converts the machine acts into text via natural language generation and generates speech according to the text via text-to-speech synthesis.

Dialogue management is the core of a SDS.  Traditionally, dialogue states are assumed to be observable and hand-crafted rules are employed for dialogue management in most commercial SDSs.  However, because of unpredictable user behaviour, inevitable ASR and SLU errors, dialogue state tracking and decision making are difficult~\citep{williams2007partially}.  Consequently, in recent years, there is a research trend from rule-based dialogue management towards statistical dialogue management.  Partially observable Markov decision process (POMDP) framework offers a well-founded theory to both dialogue state tracking and decision making in statistical dialogue management~\citep{roy2000spoken,zhang2001spoken,williams2005scaling,williams2007partially,thomson2010bayesian,gavsic2011effective,young2010hidden}.  In previous studies of POMDP, dialogue state tracking and decision making are usually investigated together.  In recent years, to advance the research of statistical dialogue management, the DST problem is raised out of the statistical dialogue management framework so that a bunch of models can be investigated for DST.

Most early studies of POMDP on DST were devoted to generative models~\citep{young2010hidden}, which learn joint probability distribution over observation and labels.  Fundamental weaknesses of generative model was revealed by the result of~\citep{williams2012challenges}.  In contrast, discriminative state tracking models have been successfully used for SDSs~\citep{deng2013recent}.  Compared to generative models where assumptions about probabilistic dependencies of features are usually needed, discriminative models directly model probability distribution of labels given observation, enabling rich features to be incorporated. The results of the Dialog State Tracking Challenge (DSTC)~\citep{williams-EtAl:2013:SIGDIAL,henderson-thomson-williams:2014:W14-43,henderson-slt14.1} further demonstrated the power of discriminative statistical models, such as Maximum Entropy (MaxEnt)~\citep{lee-eskenazi:2013:SIGDIAL}, Conditional Random Field~\citep{lee:2013:SIGDIAL}, Deep Neural Network (DNN)~\citep{sun-EtAl:2014:W14-43}, and Recurrent Neural Network (RNN)~\citep{henderson-thomson-young:2014:W14-43}.  In addition to discriminative statistical models, discriminative rule-based models have also been investigated for DST due to their efficiency, portability and interpretability and some of them showed good performance and generalisation ability in DSTC~\citep{zilka-EtAl:2013:SIGDIAL,wang-lemon:2013:SIGDIAL}.  However, both rule-based and statistical approaches have some disadvantages. Statistical approaches have shown large variation in performance and poor generalisation ability due to the lack of data~\citep{williams2012challenges}. Moreover, statistical models usually have more complex model structure and features than rule-based models, and thus can hardly achieve efficiency, portability and interpretability as rule-based models. As for rule-based models, their performance is usually not competitive to the best statistical approaches.  Additionally, since they require lots of expert knowledge and there is no general way to design rule-based models with prior knowledge, they are typically difficult to design and maintain.  Furthermore, there lacks a way to improve their performance when training data are available.

Recent studies on constrained Markov Bayesian polynomial (CMBP) framework take the first step towards bridging the gap between rule-based and statistical approaches for DST~\citep{sun-slt14,yu2014constrained}.  CMBP formulate rule-based DST in a general way and allow data-driven rules to be generated. Concretely, in the CMBP framework, DST models are defined as polynomial functions of a set of features whose coefficients are integer and satisfy a set of constraints where prior knowledge is encoded. The optimal DST model is selected by evaluating each model on training data.  \cite{yu2014constrained} further extended CMBP to real-coefficient polynomial where the real coefficients can be estimated by optimizing the DST performance on training data using grid search.  CMBP offers a way to improve the performance when training data are available and achieves competitive performance to the state-of-the-art statistical approaches, while at the same time keeping most of the advantages of rule-based models.  Nevertheless, adding features to CMBP is not as easy as to most statistical approaches because on the one hand, the features usually need to be probability related features, on the other hand, additional prior knowledge is needed to constrain the search space. For the same reason, increasing the model complexity, such as by using higher-order polynomial, by introducing hidden variables, etc. also requires additional suitable prior knowledge to be introduced to limit the search space not too large. Moreover, CMBP can hardly fully utilize the labelled data because in practice its polynomial coefficients are set by grid search.

In this paper, a novel hybrid framework, referred to as \emph{recurrent polynomial network} (RPN), is proposed to further bridge the gap between rule-based and statistical approaches for DST.  Although the basic idea for transforming rules to neural networks has been there for many years~\citep{cloete2000knowledge}, few work has been done for dialogue state tracking. RPN can be regarded as a kind of human interpretable computation networks, and its unique structure enables the framework to have all the advantages of CMBP including efficiency, portability and interpretability.  Additionally, RPN achieves more properties of statistical approaches than CMBP.  In general, RPN has neither restriction to feature type, nor search space issue to be concerned about, so adding features and increasing the model complexity are much easier in RPN. Furthermore, RPN can better explore the parameter space than CMBP with labelled data.

The DSTCs have provided the first common testbed in a standard format, along with a suite of evaluation metrics to facilitate direct comparisons among DST models~\citep{williams-EtAl:2013:SIGDIAL}. To evaluate the effectiveness of RPN for DST, both the dataset from the second Dialog State Tracking Challenge (DSTC-2) which is in restaurants domain~\citep{henderson-thomson-williams:2014:W14-43} and the dataset from the third Dialog State Tracking Challenge (DSTC-3) which is in tourists domain~\citep{henderson-slt14.1} are used. For both of the datasets, the dialogue state tracker receives SLU $N$-best hypotheses for each user turn, each hypothesis having a set of act-slot-value tuples with a confidence score.  The dialogue state tracker is supposed to output a set of distributions of the dialogue state.  In this paper, only joint goal tracking, which is the most difficult and general task of DSTC-2/3, is of interest.

The rest of the paper is organized as follows. Section \ref{sec:bridge} discusses ways of bridging rule-based and statistical approaches.  Section \ref{sec:rpn} formulates RPN.  The RPN framework for DST is described in section \ref{sec:rpnfordst}, followed by experiments in section \ref{sec:experiment}. Finally, section \ref{sec:conclusion} concludes the paper.

\section{Bridge Rule-based and Statistical Approaches}
\label{sec:bridge}
Broadly, it is straightforward to come up with two possible ways to bridge rule-based and statistical approaches: one starts from rule-based models, while the other starts from statistical models.  CMBP takes the first way, which is derived as an extension of a rule-based model \citep{sun-slt14,yu2014constrained}.  Inspired by the observation that many rule-based models such as models proposed by \cite{wang-lemon:2013:SIGDIAL} and \cite{zilka-EtAl:2013:SIGDIAL} are based on Bayes' theorem, in the CMBP framework, a DST rule is defined as a polynomial function of a set of probabilities since Bayes' theorem is essentially summation and multiplication of probabilities.  Here, the polynomial coefficients can be seen as parameters.  To make the model have good DST performance, prior knowledge or intuition is encoded to the polynomial functions by setting certain constraints to the polynomial coefficients, and the coefficients can further be optimized by data-driven optimization.  Therefore, starting from rule-based models, CMBP can directly incorporate prior knowledge or intuition into DST, while at the same time, the model is allowed to be data-driven.

More concretely, assuming that both slot and value are independent, a CMBP model can be defined as
\begin{eqnarray}
b_{t+1}(v)=&{\cal P}\left(b_t(v), P_{t+1}^+(v),P_{t+1}^-(v),\tilde{P}_{t+1}^+(v),\tilde{P}_{t+1}^-(v),1, b_t^r\right) \nonumber\\
 &{\tt s.t.}\,\,\,{\tt constraints}
 \label{eq:cmbp1}
\end{eqnarray}
where $b_t(v)$, $P_t^{+}(v)$, $P_t^{-}(v)$, $\tilde{P}_t^+(v)$, $\tilde{P}_t^-(v)$, $b_t^{r}$ are all probabilistic {\em features} which are defined as below:
\begin{itemize}
\item $b_t(v)$: belief of ``the value being $v$ at turn $t$"
\item $P_t^{+}(v)$: sum of scores of SLU hypotheses informing or affirming value $v$ at turn $t$
\item $P_t^{-}(v)$: sum of scores of SLU hypotheses denying or negating value $v$ at turn $t$
\item $\tilde{P}_t^+(v)=\sum_{v'\notin \{v,{\tt None}\}}P_t^+(v')$
\item $\tilde{P}_t^-(v)=\sum_{v'\notin \{v,{\tt None}\}}P_t^-(v')$
\item $b_t^{r}$: belief of the value being `None' (the value not mentioned) at turn $t$, i.e. $b_t^{r}=1-\sum_{v'\ne {\tt None}}b_{t}(v') $
\end{itemize}
and ${\cal P}(\cdot)$ is a multivariate polynomial function\footnote{The notation of $\sum_{0\leq k_1\leq \cdots \leq k_n\leq D}$ is shorthand for series of $n$ nested sums over ranges bounded, i.e. $\sum_{0\leq k_1\leq D}\sum_{k_1\leq k_2\leq D}\ldots\sum_{k_{n-1}\leq k_n \leq D}$.}
\begin{equation}
{\cal P}(\iota_0, \cdots, \iota_D) = \sum_{0\leq k_1\leq \cdots \leq k_n\leq D}g_{k_1,\cdots, k_n}\prod_{1\le i\le n}\iota_{k_i}
\label{eq:polynomial}
\end{equation}

where $D+1$ is the number of input variables,  $n$ is the order of the polynomial, $g_{k_1,\cdots, k_n}$ is the {\em parameter} of CMBP.
Order 3 gives good trade-off between complexity and performance, hence order 3 is used in our previous work \citep{sun-slt14,yu2014constrained} and this paper.

The {\em constraints} in equation (\ref{eq:cmbp1}) encode all necessary probabilistic conditions~\citep{yu2014constrained}. For instance,
\begin{equation}
0 \leq P_t^{+}(v) + \tilde{P}_t^+(v) \leq 1
\end{equation}
\begin{equation}
0 \leq b_t(v)\leq 1
\end{equation}  The {\em constraints} in equation (\ref{eq:cmbp1}) also encode prior knowledge or intuition~\citep{yu2014constrained}. For example, the rule {\em ``goal belief should be unchanged or positively correlated with the positive scores from SLU"} can be represented by
\begin{equation}
\frac{\partial {\cal P}\left(b_t(v), P_{t+1}^+(v), P_{t+1}^-(v), \tilde{P}_{t+1}^+(v), \tilde{P}_{t+1}^-(v), 1, b_t^r\right)}{\partial P_{t+1}^+(v)} \geq 0
\label{eq:int-constraint}
\end{equation}

The definition of CMBP formulates a search space of rule-based models, where it is easy to employ data-driven criterion to find a rule-based model with good performance.  Considering CMBP is originally motivated from Bayesian probability operation which leads to the natural use of integer polynomial coefficients ($g\in \mathbb{Z}$), the data-driven optimization can be formulated by an integer programming program~\citep{sun-slt14,yu2014constrained}.  Additionally, CMBP can also be viewed as a statistical approach. Hence, the polynomial coefficients can be extended to real numbers. The optimization of real-coefficient can be done by first getting an integer-coefficient CMBP and then performing hill climbing search~\citep{yu2014constrained}.

\section{Recurrent Polynomial Network}
\label{sec:rpn}

Recurrent polynomial network, which is proposed in this paper, takes the other way to bridge rule-based and statistical approaches.  The basic idea of RPN is to enable a kind of statistical model to take advantage of prior knowledge or intuition by using the parameters of rule-based models to initialize the parameters of statistical models.



Computational networks have been researched for decades from very basic architectures such as perceptron~\citep{rosenblatt1958perceptron} to today's various kinds of deep neural networks.  Recurrent computational networks, a class of computational networks which have recurrent connections, have also been researched for a long time, from fully recurrent network to networks with relatively complex structures such as Long short-term memory (LSTM)~\citep{hochreiter1997long}.  Like common neural networks, RPN is a statistical approach so it is as easy to add features and try complex structures in RPN as in neural networks.  However, compared with common neural networks which are ``black boxes", an RPN can essentially be seen as a polynomial function.  Hence, considering that a CMBP is also a polynomial function, the encoded prior knowledge and intuition in CMBP can be transferred to RPN by using the parameters of CMBP to initialize RPN. 
In this way, it bridges rule-based models and statistical models.

A recurrent polynomial network is a computational network. The network contains multiple edges and loops. Each node is either an {\em input node}, which is used to represent an input value, or a {\em computation node}.  Each node  $x$ is set an initial value $u_x^{(0)}$ at time $0$, and its value is updated at time $1,2,\cdots$. Both the type of edges and the type of nodes decide how the nodes' values are updated. There are two types of edges.  One type, referred to as {\em type-1}, indicates the value updating at time $t$ takes the value of a node at time $t-1$, i.e. type-1 edges are recurrent edges, while the other type, referred to as {\em type-2}, indicates the value updating at time $t$ takes another node's value at time $t$.  Except for loops made of ``type-1" edges, the network should not have loops.  For simplicity, let $I_{x}$ be the set of nodes $y$  which are linked to node  $x$ by a type-1 edge, $\hat{I}_{x}$ be the set of nodes $y$ which are linked to node  $x$ by a type-2 edge.
Based on these definitions, two types of computation nodes, {\em sum} and {\em product}, are introduced.   Specifically, at time $t>0$, if node  $x$ is a sum node, its value $u_{x}^{(t)}$ is updated by
\begin{equation}
u_{x}^{(t)}=\sum_{y \in I_{x}} w_{x,y}u_{y}^{(t-1)} + \sum_{y \in \hat{I}_{x}} \hat{w}_{x,y}u_{y}^{(t)}
\label{eq:sumnode}
\end{equation}
where $w, \hat{w} \in \mathbb{R}$ are the weights of edges. 

Similarly, if node  $x$ is a product node, its value is updated by
\begin{equation}
u_{x}^{(t)}=\prod_{y \in I_{x}} {u_{y}^{(t-1)}}^{M_{x, y}} \prod_{y \in \hat{I}_{ x}} {u_{y}^{(t)}}^{\hat{M}_{x, y}}
\label{eq:productnode}
\end{equation}
where $M_{x, y}$ and $\hat{M}_{x, y}$ are integers, denoting the multiplicity of the type-1 edge $\overrightarrow{yx}$, and the multiplicity of the type-2 edge $\overrightarrow{yx}$ respectively.  It is noted that only $w, \hat{w}$ are parameters of RPN while $M_{x,y}, \hat{M}_{x,y}$ are constant given the structure of an RPN.

\begin{figure}[htbp!]
\centering
\includegraphics[width=5cm]{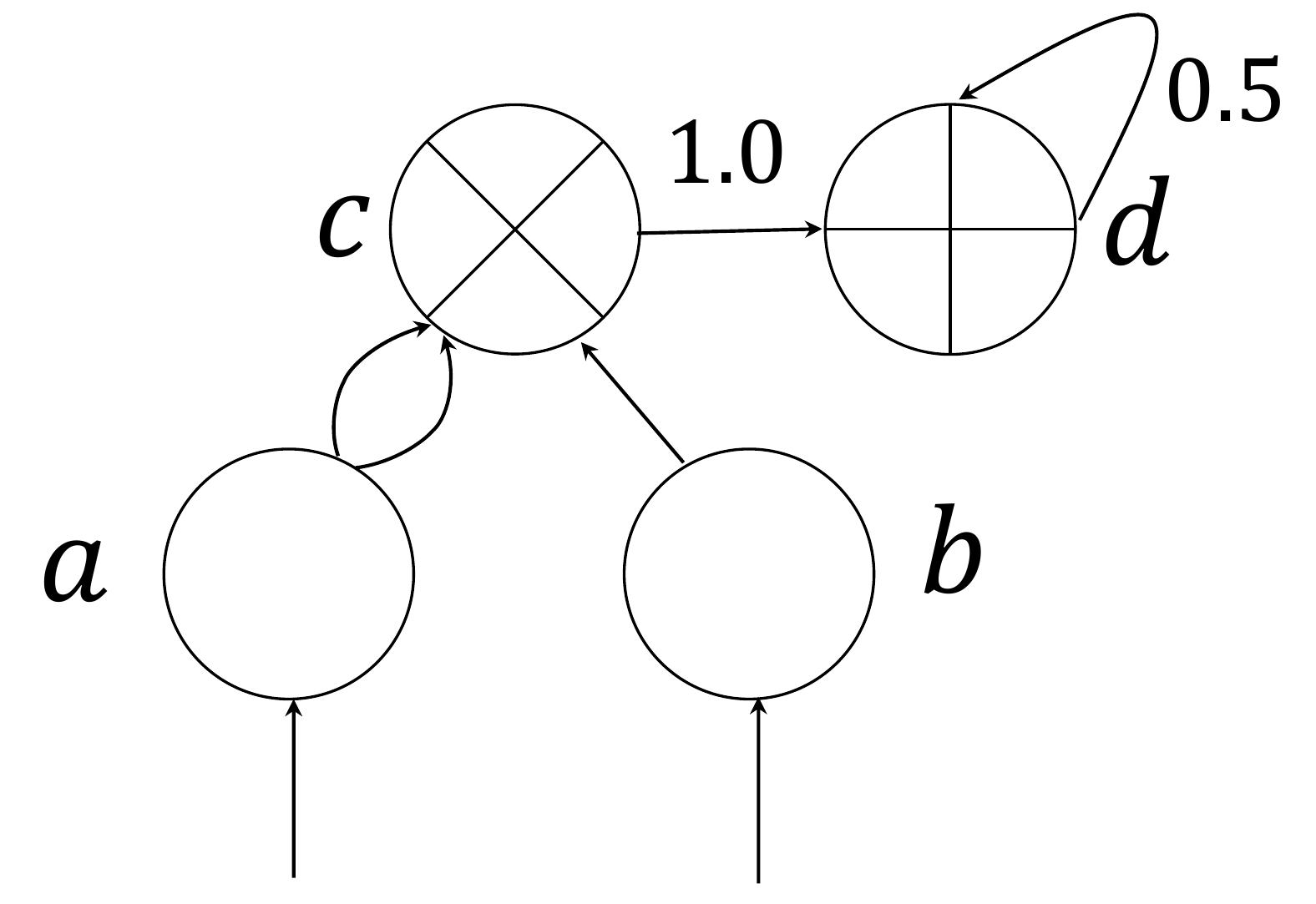}
\caption{A simple example of RPN. The type of node $a$, $b$, $c$, $d$ is input, input, product, and sum respectively. Edge $\protect\overrightarrow{dd}$ is of type-1, while the other edges are of type-2. $\hat{M}_{a,c}=2$, $\hat{M}_{b,c}=\hat{M}_{c,d}=M_{d,d}=1$.}
\label{fig:rpnexample}
\end{figure}

Let ${\bm u}^{(t)}$, $\hat{\bm {u}}^{(t)}$ denote the vector of computation nodes' values and the vector of input nodes' values at time $t$ respectively, then a well-defined RPN can be seen as a polynomial function as below.
\begin{equation}
{\bm u}^{(t)} = {\cal P}\left(\hat{\bm {u}}^{(t)} , {\bm u}^{(t-1)}, 1  \right)  
\end{equation}
where 
 ${\cal P}$ is defined by equation (\ref{eq:polynomial}). For example, for the RPN in figure \ref{fig:rpnexample}, its corresponding polynomial function is
\begin{align}
(u_c^{(t)},u_d^{(t)}) = & {\cal P}\left(u_a^{(t)},u_b^{(t)},u_c^{(t-1)},u_d^{(t-1)},1\right) \nonumber \\
= & \left({(u_a^{(t)})}^{2}u_b^{(t)}, 0.5u_d^{(t-1)}+{(u_a^{(t)})}^{2}u_b^{(t)}\right)
\end{align}

Each computation node can be regarded as an {\em output node}. For example, for the RPN in figure \ref{fig:rpnexample}, node $c$ and node $d$ can be set as output nodes.


\section{RPN for Dialogue State Tracking}
\label{sec:rpnfordst}

As introduced in section \ref{sec:introduction}, in this paper, the dialogue state tracker receives SLU $N$-best hypotheses for each user turn, each hypothesis having a set of act-slot-value tuples with a confidence score. The dialogue state tracker is supposed to output a set of distributions of the joint user goal, i.e., the value for each slot.  For simplicity and consistency with the work of \cite{sun-slt14} and \cite{yu2014constrained}, slot and value independence are assumed in the RPN model for dialogue state tracking\footnote{For DSTC-2/3 tasks, one slot can have at most one value, i.e. $0\leq \sum_{v\neq {\tt None}}b_t(v)\leq 1$.  Since value independence is assumed, to strictly maintain that relation, the belief is rescaled to ensure the sum of the belief of each value plus the belief of `None' to be $1$ when the belief is being output. Actually, to enable RPN strictly maintain that relation, our original design of RPN had a ``normalization" step when passing the belief from turn $t$ to turn $t+1$. The ``normalization" step will rescale the belief to make the sum of the belief of each value plus the belief of `None' to be 1.  Our later experiment, however, demonstrated that there was no significant performance difference between the RPN with and without the ``normalization" step. Therefore, in practice, for simplicity, the ``normalization" step can be omitted, value independence can be assumed, and the only thing needed is to rescale the belief when it is being output.}, though neither CMBP nor RPN is limited to the assumptions.  In the rest of the paper, $b_t(v), P_{t}^+(v), P_{t}^-(v), \tilde{P}_{t}^+(v), \tilde{P}_{t}^-(v)$ are abbreviated by $b_t, P_{t}^+, P_{t}^-, \tilde{P}_{t}^+, \tilde{P}_{t}^-$ respectively in circumstances where there is no ambiguity.

\subsection{Structure}
\label{subsec:structure}

\begin{figure}[htbp!]
\centering
\includegraphics[width=14cm]{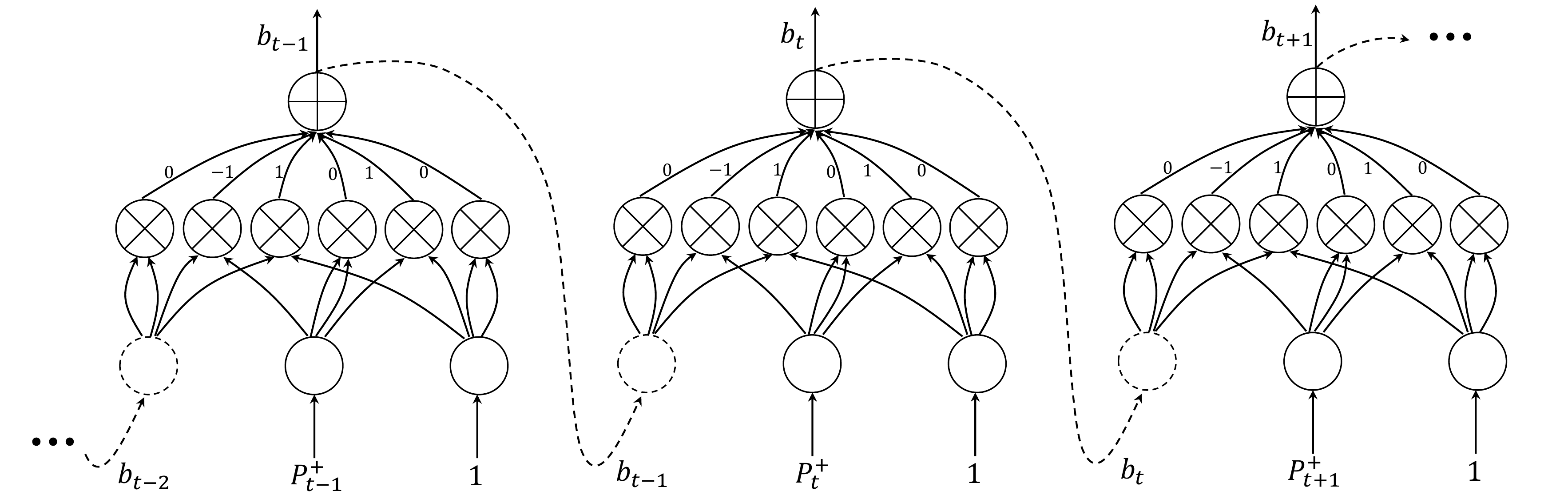}
\caption{A simple example of RPN for DST.} 
\label{fig:rpnexample2}
\end{figure}

Before describing details of the structure used in the real situations, to help understand the corresponding relationship between RPN and CMBP, let's first look at a simplified case with a smaller feature set and a smaller order, which is a corresponding relationship between the RPN shown in figure \ref{fig:rpnexample2} and 2-order polynomial (\ref{eq:simplifiedP}) with features $b_{t-1}, P^+_t, 1$:
\begin{equation}
\begin{aligned}
b_{t}&=1-(1-b_{t-1})(1-P^+_t)\\
&=b_{t-1}+P^+_t-P^+_tb_{t-1}
\end{aligned}
\label{eq:simplifiedP}
\end{equation}

Recall that a CMBP of polynomial order 2 with 3 features is the following equation (refer to equation (\ref{eq:polynomial})):

\begin{equation}
{\cal P}(\iota_0, \iota_1, \iota_2) = \sum_{0\leq k_1\leq k_2 \leq 2}g_{k_1, k_2}\prod_{1\le i\le 2}\iota_{k_i}
\label{eq:ordertwo}
\end{equation}

The RPN in figure \ref{fig:rpnexample2} has three layers. The first layer only contains input nodes. The second layer only contains product nodes. The third layer only contains sum nodes. Every product node in the second layer denotes a monomial of order 2 such as $(b_{t-1})^2, b_{t-1}, P^+_t$ and so on. Every product node in the second layer is linked to the sum node in the third layer whose value is a weighted sum of value of product nodes. With weight set according to coefficients in equation (\ref{eq:simplifiedP}), the value of sum node in the third layer is essentially the $b_t$ in equation (\ref{eq:simplifiedP}).

\begin{figure}[htbp!]
\centering
\includegraphics[width=14cm]{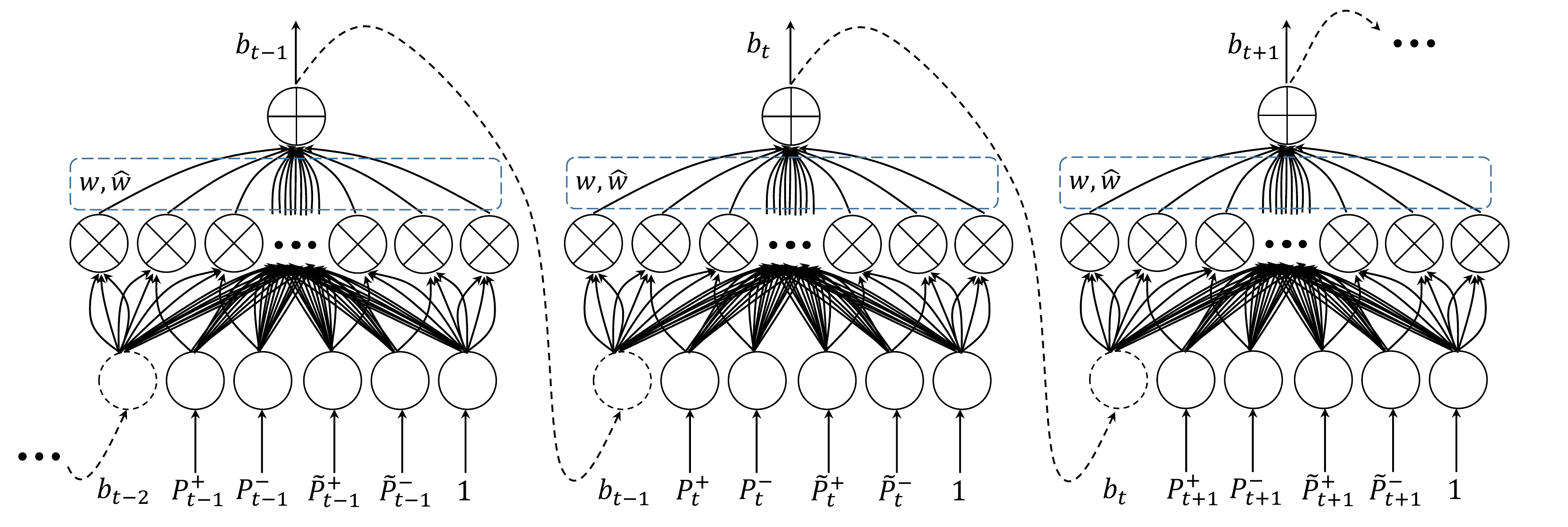}
\caption{RPN for DST. } 
\label{fig:rpnfordst}
\end{figure}

Like the above simplified case, a layered RPN structure shown in figure \ref{fig:rpnfordst} is used for dialogue state tracking in our first trial which essentially corresponds to 3-order CMBP, though the RPN framework is not limited to the layered topology. Recall that a CMBP of polynomial order 3 is used as shown in the following equation (refer to equation (\ref{eq:polynomial})):

\begin{equation}
{\cal P}(\iota_0, \cdots, \iota_D) = \sum_{0\leq k_1\leq k_2 \leq k_3\leq D}g_{k_1, k_2, k_3}\prod_{1\le i\le 3}\iota_{k_i}
\label{eq:orderthree}
\end{equation}

Let $(l,i)$ denote the index of $i$-th node in the $l$-th layer. The detailed definitions of each layer are as follows:

\begin{itemize}
\item
First layer / Input layer:

Input nodes are features at turn $t$, which corresponds to variables in CMBP in section \ref{sec:bridge}. i.e.
\begin{itemize}
\item $u_{(1, 0)}^{(t)}=b_{t-1}$
\item $u_{(1, 1)}^{(t)}=P^+_{t}$
\item $u_{(1, 2)}^{(t)}=P^-_{t}$
\item $u_{(1, 3)}^{(t)}=\tilde{P}^+_{t}$
\item $u_{(1, 4)}^{(t)}=\tilde{P}^-_{t}$
\item $u_{(1, 5)}^{(t)}=1$
\end{itemize}

While $7$ features are used in previous work of CMBP~\citep{sun-slt14,yu2014constrained}, only $6$ of them are used in RPN with feature $b_{t-1}^r$ removed\footnote{$b_{t}^r$ is the belief of value being `None', whose precise definition is given in section \ref{sec:bridge}.}. Since our experiments showed the performance of CMBP would not become worse without feature $b_{t-1}^r$, to make the structure more compact, $b_{t-1}^r$ is not used in this paper for RPN. In accordance to this, CMBP mentioned in the rest of paper does not use this feature either.


\item Second layer:

The value of every product node in the second layer is a monomial like the simplified case. And every product node has indegree 3  which is corresponding to the order of CMBP.

Every monomial in CMBP is the product of three repeatable features. Correspondingly, the value of every product node in second layer is the product of values of three repeatable nodes in the first layer. Every triple $(k_1, k_2, k_3)(0\leq k_1 \leq k_2 \leq k_3 \leq 5)$ is enumerated to create a product node $x=(2,i)$ in second layer that nodes  $(1, k_1), (1, k_2), (1, k_3)$ are linked to. i.e. $\hat{I}_x=\{(1, k_1), (1, k_2), (1, k_3) \}$.
And thus $u_x^{(t)}=u_{1, k_1}^{(t)}u_{1, k_2}^{(t)}u_{1, k_3}^{(t)}$.

And different node in the second layer is created by a distinct triple. So given the 6 input features, there are $\sum\limits_{k_1=0}^5 \sum\limits_{k_2=k_1}^5 \sum\limits_{k_3=k_2}^5 1=\binom{6 + 3 -1}{3}=56$ nodes in the second layer.

To simplify the notation, a bijection from nodes to monomials is defined as:
\begin{equation}
\begin{aligned}
{\cal F}:\{x|\mbox{$x$ is the index of a node in the $2^{nd}$ layer}\} \rightarrow \{(k_1, k_2, k_3)|0 \leq k_1 \leq k_2 \leq k_3 \leq D\}
\label{eq:bijection}
\end{aligned}
\end{equation}
\begin{equation}
{\cal F}(x)= (k_1, k_2, k_3) \Longleftrightarrow u_{2, i}^{(t)}=u_{1, k_1}^{(t)}u_{1, k_2}^{(t)}u_{1, k_3}^{(t)}
\end{equation}
where $D+1=6$ is the number of nodes in the first layer, i.e. input feature dimension.

\item Third layer:

The value of sum node  $x=(3,0)$ in the third layer is corresponding to the output value of CMBP.

Every product nodes in the second layer are linked to it. Node $x$'s value $u_{3,0}^{(t)}$ is a weighted sum of values of product node $u_{2, i}^{(t)}$ where the weights correspond to $g_{k_1, k_2, k_3}$ in equation (\ref{eq:orderthree}).

\end{itemize}

With only sum and product operation involved, every node's value is essentially a polynomial of input features. And just like recurrent neural network, node at time $t$ can be linked to node at time $t+1$.  That is why this model is called recurrent polynomial network.

The parameters of the RPN can be set 
according to CMBP coefficients $g_{k_1, k_2, k_3}$ in equation (\ref{eq:orderthree}) so that the output value is the same as the value of CMBP, which is a direct way of applying prior knowledge and intuition to statistical models. It is explained in detail in section \ref{subsec:rpninit}.

\subsection{Activation Function}
\label{subsec:activationfunc}

In DST, the output value is a belief which should lie in $[0,1]$, while values of computational nodes are not bound by certain interval in RPN.  Experiments showed that if weights are not properly set in RPN and a belief $b_{t-1}$ output by RPN is larger than 1, then $b_t$ may grow much larger because $b_t$ is the weighted sum of monomials such as $(b_{t-1})^3$. Belief of later turns such as $b_{t+10}$ will tend to infinity. 

Therefore, an activation function is needed to map $b_t$ to a legal belief value (referred to as $b'_t$) in $(0,1)$. 3 kinds of functions, the {\em logistic} function, the {\em clip} function, and the {\em softclip} function have been considered. A logistic function is defined as
\begin{equation}
logistic(x)=\frac{L}{1+e^{-\eta(x-x_0)}}
\end{equation}
It can map $\mathbb{R}$ to $(0,1)$ by setting $L=1$. However, 
since basically the RPN designed for dialogue state tracking does similar operation as CMBP which is motivated from Bayesian probability operation~\citep{yu2014constrained}, intuitively we expect the activation function to be linear on $[0,1]$ so that little distortion is added to the belief.

As an alternation, a {\em clip} function is defined as
\begin{equation}
clip(x)=
\begin{cases}
0 & \mbox{if $x < 0$} \\
x & \mbox{if $0 \leq x \leq 1$} \\
1 & \mbox{if $x > 1$} \\
\end{cases}
\end{equation}
It is linear on $[0,1]$. However, if $b'_t=clip(b_t)$, $b_t \not\in [0,1]$ and ${\cal L}$ is the loss function,
\begin{align}\frac{\partial {\cal L}}{\partial b_t}=\frac{\partial {\cal L}}{\partial b'_t} \frac{\partial b'_t}{\partial b_t}=\frac{\partial {\cal L}}{\partial b'_t}\times 0 = 0
\end{align}
Thus, $\frac{\partial {\cal L}}{\partial b_t}$ would be $0$ whatever $\frac{\partial {\cal L}}{\partial b'_t}$ is. This gradient vanishing phenomenon may affect the effectiveness of backpropagation training in section \ref{subsec:trainrpn}.

\begin{figure}[htbp!]
\centering
\includegraphics[width=15cm]{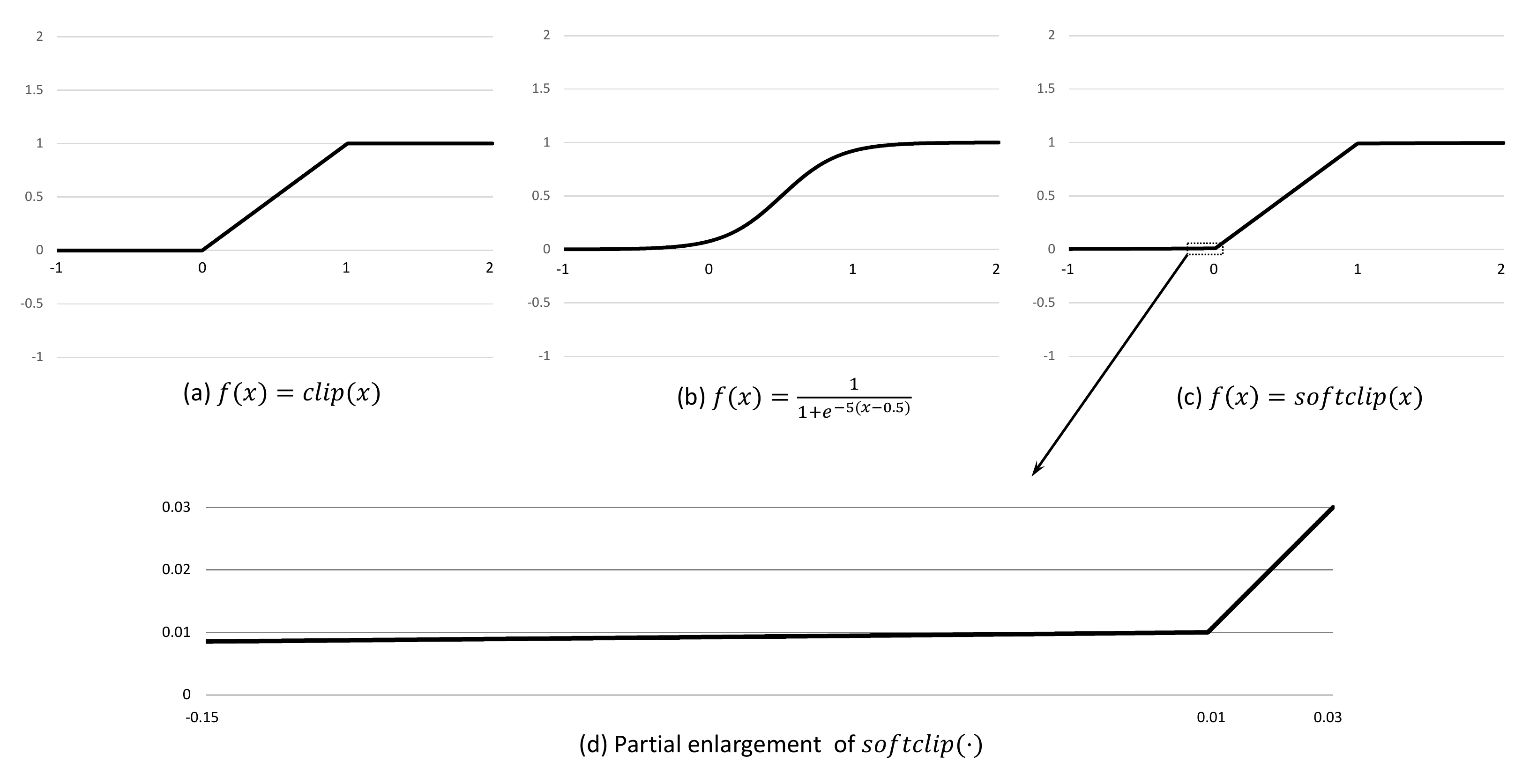}
\caption{Comparison among clip function, logistic function, and softclip function}
\label{fig:softclip}
\end{figure}

So an activation function $softclip(\cdot)$ is introduced, which is a combination of logistic function and clip function. Let $\epsilon$ denote a small value such as 0.01, $\delta$ denote the offset of sigmoid function such that $sigmoid\left(\epsilon-0.5+\delta\right) = \epsilon$. Here the {\em sigmoid} function refers to the special case of the logistic function defined by the formula
\begin{equation}
sigmoid(x)=\frac{1}{1+e^{-x}}
\end{equation}
The softclip function is defined as
\begin{equation}
softclip(x) \triangleq
\begin{cases}
sigmoid\left(x-0.5+\delta\right)& \mbox{if}\ x \leq \epsilon \\
x &\mbox{if}\ \epsilon < x < 1 - \epsilon \\
sigmoid\left(x-0.5-\delta\right) & \mbox{if}\ x \geq 1 -\epsilon\\
\end{cases}
\end{equation}

$softclip: \mathbb{R} \rightarrow (0, 1)$ is a non-decreasing, continuous function. However, It is not differentiable when $x=\epsilon$ or $x=1-\epsilon$. So we defined its derivative as follows:
\begin{equation}
\frac{\partial softclip(x)}{\partial x} \triangleq
\begin{cases}
\frac{\partial sigmoid(x-0.5+\delta)}{\partial x}& \mbox{if}\ x \leq \epsilon \\
1 & \mbox{if}\ \epsilon < x < 1 - \epsilon \\
\frac{\partial sigmoid(x-0.5-\delta)}{\partial x} &  \mbox{if}\ x \geq 1 -\epsilon\\
\end{cases}
\end{equation}

It is like a clip function. However, its derivative may be small on some inputs but is not zero. Figure \ref{fig:softclip} shows the comparison among clip function, logistic function, and softclip function.  In practice, softclip function has demonstrated better performance than both clip and logistic function\footnote{An experiment is done on the DSTC-2 dataset where RPNs using different activation functions are trained on {\tt dstc2trn} and tested on {\tt dstc2dev}. The accuracy and L2 of RPNs with $clip$, $logistic$, and $softclip$ function are (0.779, 0.329), (0.789, 0.352), (0.790, 0.317) respectively.  In particular, logistic functions with several different $\eta$ are evaluated and the result reported here is the best one.}, and was used in the rest of the paper.

With the activation function, a new type of computation node, referred to as {\em activation node}, is introduced.  Activation node only takes one input and only has one input edge of type-2, i.e $|\hat{I}_x|=1$ and $I_x=\emptyset$. The value of an activation node  $x$ is calculated as
\begin{align}
u_{x}^{(t)} &= softclip \left( u_{\jmath_x}^{(t)}\right)
 \label{eq:activation_node}
\end{align}
where $\jmath_x$ denotes the input node of node $x$. i.e. $\hat{I}_x=\{\jmath_x\}$.

\begin{figure}[htbp!]
\centering
\includegraphics[width=14cm]{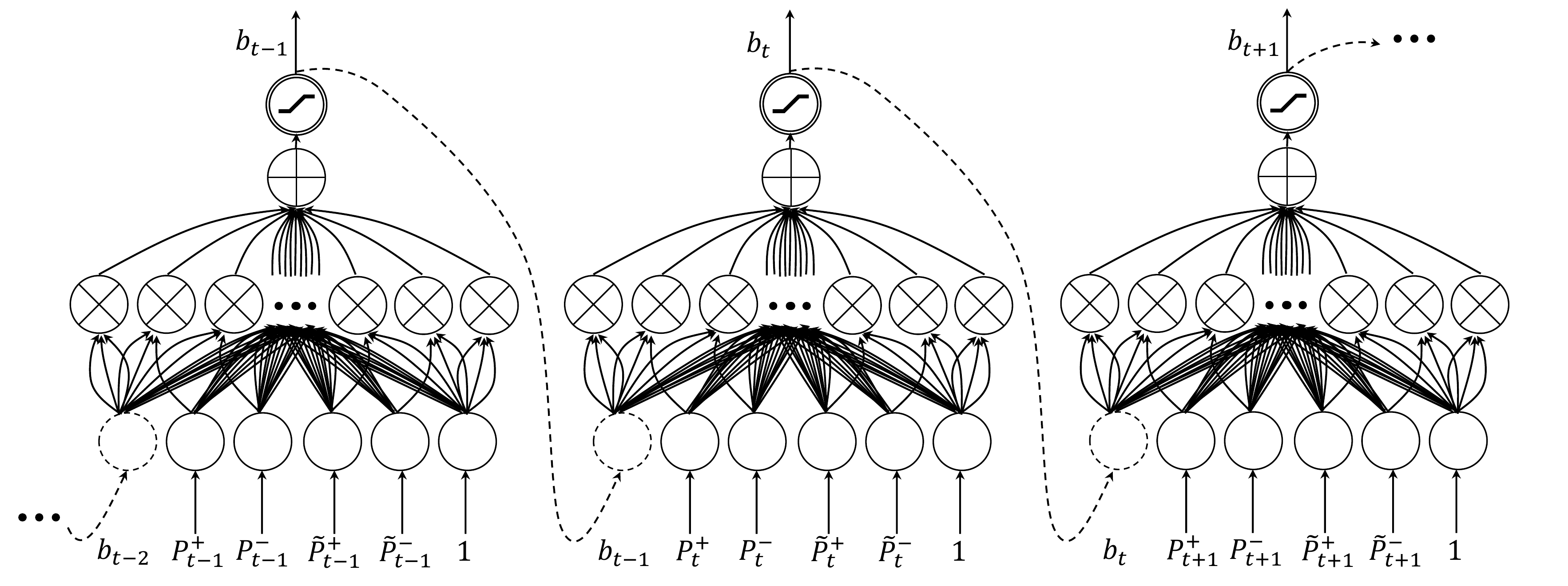}
\caption{RPN for DST with activation functions}
\label{fig:rpnfordst3}
\end{figure}

The activation function is used in the rest of the paper. Figure \ref{fig:rpnfordst3} gives an example of RPN with activation function, whose structure is constructed by adding an activation function to the RPN in figure \ref{fig:rpnfordst}. 

\subsection{Further Exploration on Structure}
\label{subsec:complexstructure}
Adding features to CMBP is not easy because additional prior knowledge is needed to add to keep the search space not too large. Concretely, adding features can introduce new monomials. Since the trivial search space is exponentially increasing as the number of monomials, the search space tends to be too large to explore when  new features are added.  Hence, to reduce the search space, additional prior knowledge is needed, which can introduce new constraints to the polynomial coefficients.
For the same reason, increasing the model complexity also requires additional suitable prior knowledge to be added to limit the search space not too large in CMBP.

In contrast to that, since RPN can be seen as a statistically model, it is as easy as most statistical approaches such as RNN to add new features to RPN and use more complex structures. At the same time, no matter what new features are used and how complex the structure is, RPN can always take advantage prior knowledge and intuition which is discussed in section \ref{subsec:rpninit}.  In this paper, both new features and complex structure are explored.

Adding new features can be done by just adding input nodes which correspond to the new features, and then adding product nodes corresponding to the new possible monomials introduced by the new features.  In this paper, for slot $s$, value $v$ at turn $t$, in addition to $f_0 \sim f_5$ which are defined as $b_{t-1}(v)$, $P_t^{+}(v)$, $P_t^{-}(v)$, $\tilde{P}_t^+(v)$, $\tilde{P}_t^-(v)$, and $1$ respectively, 4 new features are investigated.
$f_6$ and $f_7$ are features of system acts at the last turn: for slot $s$, value $v$ at turn $t$,
\begin{itemize}
\item $f_6 \triangleq canthelp(s, t, v) \cup canthelp.missing\_slot\_value(s,t)$ =$1$ if the system cannot offer a venue with the constraint $s=v$ or the value of slot $s$ is not known for the selected venue, otherwise $0$.
\item $f_7 \triangleq select(s, t, v)$  =$1$ if the system asks the user to pick a suggested value for slot $s$, otherwise $0$.
\end{itemize}
$f_6$ and $f_7$ are introduced because user is likely to change their goal if given machine acts $canthelp(s,t,v)$, $canthelp.missing\_slot\_value(s,t)$ and $select(s,t,v)$.
$f_8$ and $f_9$ are features of user acts at the current turn: for slot $s$, value $v$ at turn $t$,
\begin{itemize}
\item $f_8 \triangleq inform(s, t, v)$ = $1$ if one of SLU hypotheses from the user is informing slot $s$ is $v$, otherwise $0$.
\item $f_9 \triangleq deny(s, t, v) $ =$1$ if one of SLU hypotheses from the user is denying slot $s$ is $v$, otherwise $0$.
\end{itemize}
$f_8$ and $f_9$ are features about SLU acttype, introduced to make system robust when the confidence scores of SLU hypothesis are not reliable.

In this paper, the complexity of evaluating and training RPN for DST would not increase sharply because a constant order 3 is used and number of product nodes in the second layer grows from 56 to 220 when number of features grows from 6 to 10.

In addition to new features, RPN of more complex structure is also investigated in this paper. To capture some property just like belief $b_t$ of dialogue process, a new sum node  $x=(3, 1)$ in the third layer is introduced. The connection of $(3,1)$ is the same as $(3,0)$, so it introduces a new recurrent connection. The exact meaning of its value is unknown. However, it is the only value used to record information other than $b_t$ of previous turns. Every other input features except $b_t$ are features of current turn $t$. Compared with $b_t$, there are fewer restrictions on the value of $(3, 1)$ since its value is not directly supervised by the label. Hence, introducing $(3,1)$ may help to reduce the effect of inaccurate labels.

\begin{figure}[htbp!]
\centering
\includegraphics[width=14cm]{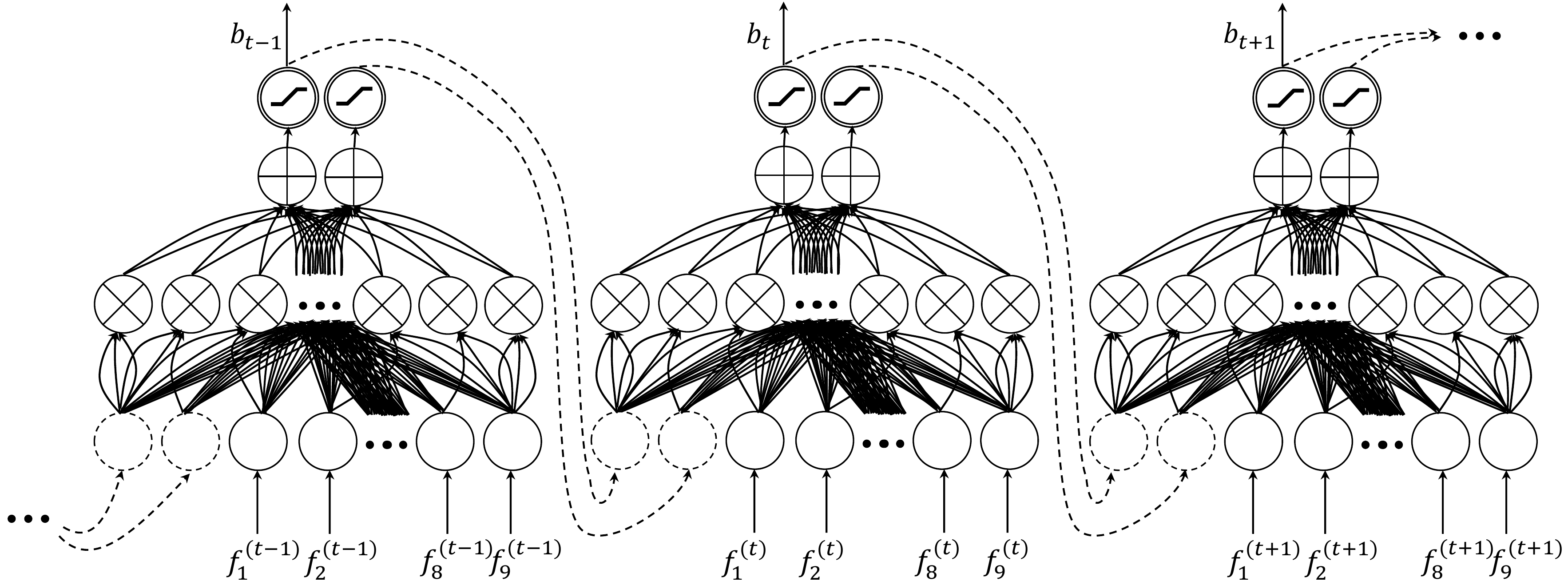}
\caption{RPN with new features and more complex structure for DST}
\label{fig:rpnfordst2}
\end{figure}

The structure of the RPN with 4 new features and 1 new sum node, together with new activation nodes introduced in section \ref{subsec:activationfunc} is shown in figure \ref{fig:rpnfordst2}.

\subsection{RPN Initialization}
\label{subsec:rpninit}

Like most neural network models such as RNN, the initialization of RPN can be done by setting each weight, i.e. $w$ and $\hat{w}$, to be a small random value. However, with its unique structure, the initialization can be much better by taking advantage of the relationship between CMBP and RPN which is introduced in section \ref{subsec:structure}.

When RPN is initialized according to a CMBP, prior knowledge and constraints are used to set RPN's initial parameters as a suboptimum point in the whole parameter space. RPN as a statistical model can fully utilize the advantages of statistical approaches. However, RPN is better than real CMBP while they both use data samples to train parameters. In the work of \cite{yu2014constrained}, real-coefficient CMBP uses hill climbing to adjust parameters that are initially not zero and the change of parameters are always a multiple of 0.1. RPN can adjust all parameters including parameters initialized as 0 concurrently, while the complexity of adjusting all parameters concurrently is nearly the same as adjusting one parameter in CMBP. Besides, the change of parameters can be large or small, depending on learning rate. Thus, RPN and CMBP both are bridging rule-based models and statistical ones, while RPN is a statistical model utilizing rule advantages and CMBP is a rule model utilizing statistical advantages.


In fact, given a CMBP, an RPN can achieve the same performance as the CMBP just by setting its weights 
according to the coefficients of the CMBP. To illustrate that, the steps of initializing the RPN in figure \ref{fig:rpnfordst2} with a CMBP of features $f_0 \sim f_9$ is described below.

First, to ensure that the new added sum node  $x=(3, 1)$ will not influence the output $b_t$ in RPN with initial parameters, $\hat{w}_{x,y}$ is set to $0$ for all $y$. So node  $x$'s value $u_x^{(t)}$ is always $0$.

Next, considering the RPN in figure \ref{fig:rpnfordst2} has more features than CMBP does, the weights related the new features should be set to 0. Specifically, suppose node  $x$ is the sum node in the third layer in RPN denoting $b_t$ before activation and node  $y$ is one of the product nodes in the second layer denoting a monomial, if product node  $y$ is products of features $f_6, f_7, f_8, f_9$ or the added sum node, then node  $y$'s value is not a monomial in CMBP, then weights $\hat{w}_{x,y}$ should be set to $0$.

Finally, if product node  $y$ is the product of features $f_0 \sim f_5$, suppose the order of CMBP is 3, then ${\cal F}(y)=(k_1, k_2, k_3)$ defined in equation (\ref{eq:bijection}) should satisfy $0 \leq k_1 \leq k_2 \leq k_3 \leq 5$. Weights $\hat{w}_{xy}$ should be initialized as $g_{k_1, k_2, k_3}$ which is the coefficient of $f_{k_1}f_{k_2}f_{k_3}$ in CMBP. Thus,

\begin{equation}
w_{x, y} =
\begin{cases}
g_{k_1, k_2, k_3} &   \mbox{if $x=(2, 0)$ and ${\cal F}(x)=(k_1, k_2, k_3)$}\\
0 &  \mbox{otherwise}
\end{cases}
\end{equation}

For RPN of other structures, the initialization can be done by following similar steps.

Experiments show that after training, there are only a few weights larger than 0.1, no matter using CMBP or random initialization.

\subsection{Training RPN}
\label{subsec:trainrpn}


Suppose $T(d)$ is the number of turns in dialogue $d$, $H(d, s, t)$ is the set of values corresponding to slot $s$ appearing in SLU hypothesis in turn $t$ in dialogue $d$, $b_{d, s, t}(v)$ is the output belief of value $v$ in dialogue $d$ and $l_{d, s,t}(v)$ is the indicator of goal $s=v$ being part of joint goal at turn $t$ in the label of dialogue $d$. The cost function is defined as

\begin{align}
{\cal L} = \frac{1}{\sum_{d} T(d)} \sum_{d} \sum_{t=0}^{T(d)-1} \sum_{s} \sum_{v \in \cup_{i=0}^t H(d, s, i)} (b_{d, s, t}(v) - l_{d, s, t}(v))^2
\end{align}

\AlgoDontDisplayBlockMarkers\SetAlgoNoEnd

\begin{algorithm}
\caption{Training Algorithm of RPN for DST}
\label{alg:training}
\ForEach {Mini batch $m$}
{
Initialize $\Delta w_{xy}=0, \Delta \hat{w}_{x,y} = 0$ for every $x,y$

Initialize the value of recurrent node at turn 0 as 0

\ForEach{Training dialogue $d$, slot $s$, value $v$ in mini batch $m$}
{
  $T \leftarrow$ the number of turns of current training sample
  
  \tcc{forward pass}
  \For {$t\leftarrow 1$ \KwTo $T$} 
  {
        \For {$d\leftarrow 1$ \KwTo $4$} 
        {
        	\ForEach{node  $x$ in time $t$, layer $d$}
        	{      	        	
        	  evaluate $u_x^{(t)}$
        	    	
        	  \If{$x$ is output node}
        	  {
        	  $\delta_x^{(t)} \leftarrow 2(u_x^{(t)}-l_t)$
        	  }
        	  \Else
        	  {
        	  $\delta_x^{(t)} \leftarrow 0$
        	  }
        	}
        }
  }
    \tcc{backward pass}
    \For {$t\leftarrow T$ \KwTo $1$}
  {
        \For {$d\leftarrow 4$ \KwTo $1$} 
        {
        	\ForEach{node  $x$ in time $t$, layer $d$}
        	{
        		\ForEach{node $y \in \hat{I}_x $} 
        		{
        			\tcc{Node $y$ is linked to node $x$ by a "type-2" edge}        		
        			$\delta_y^{(t)} \leftarrow \delta_y^{(t)} + \delta_x^{(t)}\frac{\partial u_x^{(t)}}{\partial u_y^{(t)}}$ 
        			
        		}
	       		\ForEach{node $y \in I_x $}
        		{
        			\tcc{Node $y$ is linked to node $x$ by a "type-1" edge}
        			$\delta_y^{(t-1)} \leftarrow \delta_y^{(t-1)} + \delta_x^{(t)}\frac{\partial u_x^{(t)}}{\partial u_y^{(t-1)}}$
        		}
        	}        	
        }
  }
      \For {$t\leftarrow 1$ \KwTo $T$}
  {
        \For {$d\leftarrow 1$ \KwTo $4$} 
        {
        	\ForEach{sum node  $x$ in time $t$, layer $d$}
        	{
        		\ForEach{node $y \in \hat{I}_x $}
        		{
					$\Delta \hat{w}_{xy} \leftarrow \Delta \hat{w}_{xy} +\alpha \delta_x^{(t)} u_y^{(t)}$
        		}
	       		\ForEach{node $y \in {I}_x $}
        		{
					$\Delta {w}_{xy} \leftarrow \Delta {w}_{xy} +\alpha \delta_x^{(t)} u_y^{(t-1)}$
				}
        	}        	
        }
  }
}
\ForEach {edge$(x,y)$}
{
$w_{xy} \leftarrow w_{xy} - \Delta w_{xy} $

$\hat{w}_{xy} \leftarrow \hat{w}_{xy} - \Delta \hat{w}_{xy} $
}
}
\end{algorithm}

Training process of a mini-batch can be divided into two parts: forward pass and backward pass.

\paragraph{Forward Pass}
For each training sample, every node's value at every time is evaluated first. When evaluating $u_{x}^{(t)}$, values of nodes in $I_x$ and $\hat{I}_x$ should be evaluated before. The computation formula should be based on the type of node $x$. In particular, for a layered RPN structure, we can simply evaluate $u_{x_1}^{t_1}$ earlier than $u_{x_2}^{t_2}$ if $t_1 < t_2$ or $t_1=t_2$ and $x_1$'s layer number is smaller than $x_2$'s.

\paragraph{Backward Pass}

Backpropagation through time (BPTT) is used in training RPN.
Let error of node  $x$ at time $t$ be $\delta_x^{(t)} = \frac{\partial {\cal L}}{\partial u_x^{(t)}}$. If a node  $x$ is an output node, then $\delta_x^{(t)}$ should be initialized according to its label $l_t$ and output value $u_x^{(t)}$, otherwise $\delta_x^{(t)}$ should be initialized to $0$. 
After a node's error $\delta_x^{(t)}$ is determined, it can be passed to $\delta_y^{(t-1)} (y \in I_x)$ and $\delta_y^{(t)} (y \in \hat{I}_x)$.
Error passing should follow the reversed edge's direction. So the order of nodes passing error can follow the reverse order of evaluating nodes' values.

When every $\delta_x^{(t)}$ has been evaluated, the increment on weight $\hat{w}_{xy}$ can be calculated by
\begin{equation}
\begin{aligned}
\Delta \hat{w}_{xy} &=\alpha \frac{\partial {\cal L}}{\partial \hat{w}_{xy}}\\
&=\alpha \sum_{i=1}^T  \frac{\partial {\cal L}}{\partial u_x^{(t)}} \frac{\partial u_x^{(t)}}{\partial \hat{w}_{xy}}\\
&=\alpha \sum_{i=1}^T \delta_x^{(t)} u_y^{(t)}
\end{aligned}
\end{equation}

where $\alpha$ is the learning rate. $\Delta w_{xy}$ can be evaluated similarly.

Note that only $ w_{xy}$ and $\hat{w}_{xy}$ are parameters of RPN.

The complete formula of evaluating node value $u_{x}^{(t)}$ and passing error $\delta_x^{(t)}$ can be found in appendix.

In this paper, mini-batch is used in training RPN for DST with batch size $8$. In each training epoch, $\Delta w_{xy}$ and $\Delta \hat{w}_{xy}$ are calculated for every training sample and added together. The weight $w_{xy}$ and $\hat{w}_{xy}$ is updated by
\begin{align}
w_{xy} &= w_{xy} - \Delta w_{xy} \\
\hat{w}_{xy} &= \hat{w}_{xy} - \Delta \hat{w}_{xy}
\end{align}

The pseudocode of training is shown in algorithm \ref{alg:training}.

\section{Experiment}
\label{sec:experiment}

As introduced in section \ref{sec:introduction}, in this paper, DSTC-2 and DSTC-3 tasks are used to evaluate the proposed approach. Both tasks provide training dialogues with turn-level ASR hypotheses, SLU hypotheses and user goal labels. The DSTC-2 task provides 2118 training dialogues in restaurants domain~\citep{henderson-thomson-williams:2014:W14-43}, while in DSTC-3, only 10 in-domain training dialogues in tourists domain are provided, because the DSTC-3 task is to adapt the tracker trained on DSTC-2 data to the new domain with very few dialogues~\citep{henderson-slt14.1}. Table \ref{tab:datacondition} summarizes the size of datasets of DSTC-2 and DSTC-3.
\begin{table}[htbp!]
\centering
\begin{tabular}{|c|c||c|c|}
\hline
Task & Dataset & \#Dialogues & Usage\\
\hline\hline
&{\tt dstc2trn}  & 1612 & Training\\
DSTC-2 & {\tt dstc2dev} & 506 & Training\\
&{\tt dstc2eval} & 1117 & Test\\
\hline
\multirow{2}*{DSTC-3} & {\tt dstc3seed} & 10 & Not used\\
& {\tt dstc3eval} & 2265 & Test\\
\hline
\end{tabular}
\caption{Summary of data corpora of DSTC-2/3\label{tab:datacondition}}
\end{table}

The DST evaluation criteria are the joint goal {\em accuracy} and the {\em L2}~\citep{henderson-thomson-williams:2014:W14-43,henderson-slt14.1}. {\em Accuracy} is defined as the fraction of turns in which the tracker's 1-best joint goal hypothesis is correct, the larger the better. {\em L2} is the L2 norm between the distribution of all hypotheses output by the tracker and the correct goal distribution (a delta function), the smaller the better. Besides, schedule 2 and labelling scheme A defined in \citep{henderson2013dialog} are used in both tasks.  Specifically, schedule 2 only counts the turns where new information about some slots either in a system confirmation action or in the SLU list is observed. Labelling scheme A is that the labelled state is accumulated forwards through the whole dialogue. For example, the goal for slot $s$ is ``{\tt None}" until it is informed as $s=v$ by the user, from then on, it is labelled as $v$ until it is again informed otherwise.

It has been shown that the organiser-provided live SLU confidence was not good enough~\citep{zhu-slt14,sun-EtAl:2014:W14-43}. Hence, most of the state-of-the-art results from DSTC-2 and DSTC-3 used refined SLU (either explicitly rebuild a SLU component or take the ASR hypotheses into the trackers~\citep{williams:2014:W14-43,sun-EtAl:2014:W14-43,henderson-thomson-young:2014:W14-43,henderson-slt14.2,kadlec-slt14,sun-slt14}). In accordance to this, except for the results directly taken from other papers (shown in table \ref{tab:performancedstc2} and \ref{tab:performancedstc3}), all experiments in this paper used the output from a refined semantic parser~\citep{zhu-slt14,sun-EtAl:2014:W14-43} instead of the live SLU provided by the organizer.

For all experiments, MSE\footnote{MSE is chosen because of two reasons: (i) MSE can directly reflect L2 performance which is one of the main metrics in DSTCs. (ii) Experiment has shown that some other criterion such as cross-entropy loss cannot lead to better performance.} is used as the training criterion and full-batch batch is used.  For both DSTC-2 and DSTC-3 tasks, {\tt dstc2trn} and {\tt dstc2dev} are used, 60\% of the data is used for training and 40\% for validation, unless otherwise stated.  Validation is performed every 5 epochs. Learning rate is set to 0.6 initially. During the training, learning rate is halved each time the performance does not increases. 
Training is stopped when the learning rate is sufficiently small, or the maximum number of training epochs is reached.  Here, the maximum number of training epochs is set to 40.  L2 regularization is used for all the experiments\footnote{The parameter of L2 regularization is set to be the one leading to the best performance on {\tt dstc2dev} when trained on {\tt dstc2trn}. L1 regularization is not used since it cannot yield better performance.}.

\subsection{Investigation on RPN Configurations}
\label{subsec:rpnconfig}
This section describes the experiments comparing different configurations of RPN. All experiments were performed on both the DSTC-2 and DSTC-3 tasks.

As indicated in section \ref{subsec:rpninit}, an RPN can be initialized by a CMBP. Table \ref{tab:diffinit} shows the performance comparison between initialization with a CMBP and with random values. In this experiment, the structure shown in figure \ref{fig:rpnfordst3} is used.  The performance of initialization with random values\footnote{The random scheme used is the one with the best performance on {\tt dstc2dev} when trained on {\tt dstc2trn} among various kinds of random initialization scheme.} reported here is the average performance of 10 different random seeds, and their standard deviations are given in parentheses.  


\begin{table}[htbp!]
\centering
\begin{tabular}{|c||c|c|c|c|}
\hline
\multirow{2}*{Initialization} & \multicolumn{2}{|c|}{{\tt dstc2eval}} & \multicolumn{2}{|c|}{{\tt dstc3eval}} \\
\cline{2-5}
  & \tt Acc & \tt L2 & \tt Acc & \tt L2  \\
\hline\hline
 Random   & 0.753 (0.008) & 0.468 (0.020) & 0.633 (0.005) & 0.667 (0.026) \\ 
\hline
 CMBP   & 0.756 & 0.373 & 0.648 & 0.553 \\ 
\hline

\end{tabular}
\caption{Performance comparison between the RPN initialized by random values, and the RPN initialized by the CMBP coefficients on {\tt dstc2eval} and {\tt dstc3eval}. \label{tab:diffinit}}
\end{table}

The performance of the RPN initialized by random values 
is compared with the performance of the RPN initialized by the integer-coefficient CMBP. Here, the CMBP has $12$ non-zero coefficients and has the best performance on {\tt dstc2dev} when trained on {\tt dstc2trn}. 
It can be seen from table \ref{tab:diffinit} that the RPN initialized by the CMBP coefficients outperforms the RPN initialized by random values moderately on {\tt dstc2eval} and significantly on {\tt dstc3eval} (p-value$ < 0.05$). This demonstrates the encoded prior knowledge and intuition in CMBP can be transferred to RPN to improve RPN's performance, which is one of RPN's advantage,
bridging rule-based models and statistical models.
In the rest of the experiments, all RPNs use CMBP coefficients for initialization.

Since section \ref{subsec:complexstructure} shows that it is convenient to add features and try more complex structures, it is interesting to investigate RPNs with different feature sets and structures, as shown in table \ref{tab:diffconfig}.  It can be seen that while no obvious correlation between the performance and different configurations of feature sets and structures can be observed on {\tt dstc2eval} and {\tt dstc3eval}, RPNs with new features and new recurrent connections have achieved slightly better performance
.  Thus, in the rest of the paper, both new features and new recurrent connections are used in RPN, unless otherwise stated.

\begin{table}[htbp!]
\centering
\begin{tabular}{|c|c||c|c|c|c|}
\hline
\multirow{2}*{Feature Set} & New Recurrent & \multicolumn{2}{|c|}{{\tt dstc2eval}} & \multicolumn{2}{|c|}{{\tt dstc3eval}} \\
\cline{3-6}
& Connections & \tt Acc & \tt L2 & \tt Acc & \tt L2  \\
\hline\hline
 $f_0 \sim f_5$  &  \multirow{2}*{No}  & 0.756 & 0.373 &0.648 &0.553 \\ 
 \cline{1-1} \cline{3-6}
 $f_0 \sim f_9$  &   &  0.757 & 0.374  & 0.650  & 0.557 \\ 
\hline
 $f_0 \sim f_5$  &  \multirow{2}*{Yes}  & 0.756 & 0.373 & 0.648& 0.553\\ 
 \cline{1-1} \cline{3-6}
 $f_0 \sim f_9$  &   &0.757 &0.374 & 0.650 & 0.549 \\ 
\hline
\end{tabular}
\caption{Performance comparison among RPNs with different configurations on {\tt dstc2eval} and {\tt dstc3eval}. \label{tab:diffconfig}}
\end{table}


\subsection{Comparison with Other DST Approaches}
\label{subsec:cmpwithotherdst}

The previous subsection investigates how to get the RPN with the best configuration.  In this subsection, the performance of RPN is compared to both rule-based and statistical approaches. To make fair comparison, all statistical models together with RPN in this subsection use similar feature set.  Altogether, 2 rule-based trackers and 3 statistical trackers were built for performance comparison.

\begin{table}[htbp!]
\centering
\begin{tabular}{|c|c||c|c|c|c|}
\hline
\multirow{2}*{Type} & \multirow{2}*{System} & \multicolumn{2}{|c|}{{\tt dstc2eval}} & \multicolumn{2}{|c|}{{\tt dstc3eval}} \\ \cline{3-6}
   &  & \tt Acc & \tt L2 & \tt Acc & \tt L2  \\
\hline\hline
\multirow{2}*{Rule} & MaxConf & 0.668 & 0.647 & 0.548 & 0.861 \\
& HWU & 0.720  & 0.445  & 0.594  & 0.570  \\
\hline
\multirow{2}*{Statistical} & DNN & 0.719 & 0.469 & 0.628 & 0.556 \\
& MaxEnt & 0.710 &  0.431  & 0.607 & 0.563 \\
& LSTM &0.736  & 0.418  & 0.632  & 0.549 \\
\hline
\multirow{2}*{Mixed} & CMBP & 0.755 & 0.372 &0.627 & 0.546 \\ 
& RPN & 0.756 & 0.373 & 0.648 & 0.553 \\ 
\hline
\end{tabular}
\caption{Performance comparison among RPN, rule-based and statistical approaches with similar feature set on {\tt dstc2eval} and {\tt dstc3eval}. The performance of CMBP in the table is the performance of the RPN which has been initialized but not been trained.  \label{tab:vswithsimiliarfeature}}
\end{table}

\begin{itemize}
\item \textbf{MaxConf} is a {\em rule-based} model commonly used in spoken dialogue systems which always selects the value with the highest confidence score from the $1^{st}$ turn to the current turn. It was used as one of the primary baselines in DSTC-2 and DSTC-3.
\item \textbf{HWU} is a {\em rule-based} model proposed by \cite{wang-lemon:2013:SIGDIAL}. It is regarded as a simple, yet competitive baseline of DSTC-2 and DSTC-3.
\item \textbf{DNN} is a {\em statistical model} using deep neural network model~\citep{sun-EtAl:2014:W14-43}
 with probability feature as RPN. Since DNN does not have recurrent structures while RPN does, to fairly take into account this, the DNN feature set at the $t^{th}$ turn is defined as
\begin{equation}
\bigcup_{i\in \{t-9,\cdots,t\}}\left\{P_{i}^+(v),P_{i}^-(v),\tilde{P}_{i}^+(v),\tilde{P}_{i}^-(v)\right\}\cup \left\{\hat{P}(t)\right\}\nonumber
\end{equation}
where $\hat{P}(t)$ is the highest confidence score from the $1^{st}$ turn to the $t^{th}$ turn. The DNN has 3 hidden layers with 64 nodes per layer.
\item \textbf{MaxEnt} is also a {\em statistical model} using Maximum Entropy model~\citep{sun-EtAl:2014:W14-43}
 with the same input features as DNN.
\item \textbf{LSTM} is another {\em statistical model} using long short-term memory model~\citep{hochreiter1997long} with the same input features as RPN. It has similar structure as the DNN model~\citep{sun-EtAl:2014:W14-43} except for its hidden layers using LSTM blocks.  The LSTM used here has 3 hidden layers with 100 LSTM blocks per layer.  
\end{itemize}

It can be observed that, with similar feature set, RPN can outperform both rule-based and statistical approaches in terms of joint goal accuracy. Statistical significance tests were also performed assuming a binomial distribution for each turn. RPN was shown to significantly outperform both rule-based and statistical approaches at 95\% confidence level. For L2, RPN is competitive to both rule-based and the statistical approaches. 

\subsection{Comparison with State-of-the-art DSTC Trackers}

In the DSTCs, the state-of-the-art trackers mostly employed statistical approaches. Usually, richer feature set and more complicated model structures than the statistical models in section \ref{subsec:cmpwithotherdst} are used.  In this section, the proposed RPN approach is compared to the best submitted trackers in DSTC-2/3 and the best CMBP trackers, regardless of fairness of feature selection and the SLU refinement approach. RPN is compared and the results are shown in table \ref{tab:performancedstc2} and table \ref{tab:performancedstc3}. Note that structure shown in figure \ref{fig:rpnfordst2} with richer feature set and a new recurrent connection is used here.

\begin{table}[htbp!]
\centering
\begin{tabular}{|c|c||c|c|c|}
\hline
System   & Approach & Rank & Acc & L2 \\
\hline\hline
Baseline* & Rule & 5 & 0.719 & 0.464 \\
\hline
\cite{williams:2014:W14-43} & LambdaMART& 1 & {0.784}  & {0.735}  \\
\cite{henderson-thomson-young:2014:W14-43}& RNN & 2  & 0.768 & 0.346   \\
\cite{sun-EtAl:2014:W14-43} & DNN  & 3 & 0.750 & 0.416  \\
\hline
\cite{yu2014constrained} & Real CMBP & 2.5 & 0.762 & 0.436 \\
\hline
RPN & RPN & 2.5 & 0.757 & 0.374 \\ 
\hline
\end{tabular}
\caption{Performance comparison among RPN, real-coefficient CMBP and best trackers of DSTC-2 on {\tt dstc2eval}. Baseline* is the best results from the 4 baselines in DSTC2.\label{tab:performancedstc2}}
\end{table}

Note that, in DSTC-2, the \cite{williams:2014:W14-43}'s system employed batch ASR hypothesis information (i.e. off-line ASR re-decoded results) and cannot be used in the normal on-line model in practice. Hence, the practically best tracker is \cite{henderson-thomson-young:2014:W14-43}.  It can be observed from table \ref{tab:performancedstc2}, 
RPN ranks only second to the best practical tracker among the submitted trackers in DSTC-2 in accuracy and L2. 
 Considering that RPN only used probabilistic features and very limited added features and can operate very efficiently, it is quite competitive.

\begin{table}[htbp!]
\centering
\begin{tabular}{|c|c||c|c|c|}
\hline
 System  & Approach & Rank & Acc & L2  \\
\hline\hline
Baseline* & Rule & 6 & 0.575 & 0.691\\
\hline
\cite{henderson-slt14.2} & RNN & 1 & 0.646 & 0.538 \\
\cite{kadlec-slt14} & Rule & 2 & 0.630 & 0.627 \\
\cite{sun-slt14} & Int CMBP  & 3 & 0.610 & 0.556 \\
\hline
\cite{yu2014constrained} & Real CMBP & 1.5 & 0.634 & 0.579 \\
\hline
RPN & RPN &  0.5 & 0.650  & 0.549 \\ 
\hline
\end{tabular}
\caption{Performance comparison among RPN, real-coefficient CMBP and best trackers of DSTC-3 on {\tt dstc3eval}. Baseline* is the best results from the 4 baselines in DSTC3.\label{tab:performancedstc3}}
\end{table}

It can be seen from table \ref{tab:performancedstc3}, RPN trained on DSTC-2 can achieve state-of-the-art performance on DSTC-3 without modifying tracking method\footnote{The parser is refined for DSTC-3~\citep{zhu-slt14}.}, outperforming all the submitted trackers in DSTC-3 including the RNN system.
  This demonstrates that RPN successfully inherits the advantage of good generalization ability of rule-based model. Considering the feature set and structure of RPN are relatively simple in this paper, future work will investigate richer features and more complex structures.

\section{Conclusion}
\label{sec:conclusion}
This paper proposes a novel hybrid framework, referred to as recurrent polynomial network, to bridge the rule-based model and statistical approaches. With the ability of incorporating prior knowledge into a statistical framework, RPN has the advantages of both rule-based and statistical approaches.  
Experiments on two DSTC tasks showed that the proposed approach not only is more stable than many major statistical approaches, but also has competitive performance, outperforming many state-of-the-art trackers.

Since the RPN in this paper only used probabilistic features and very limited added features, the performance of RPN can be influenced by how reliable the SLU's confidence scores are.  Therefore, future work will investigate the influence of SLUs on the performance of RPN, and rich features for RPN.   Moreover, future work will also address applying RPN to other domains, such as the bus timetables domain in DSTC-1, and theoretic analysis of RPN.


\bibliography{ref}

\section*{Appendix}
\subsection*{Derivative calculation}
\label{subsec:appen}
Using MSE as the criterion, $\delta_{x}^{(t)}=\frac{\partial {\cal L}}{\partial u_x^{(t)}}$ is initialized as following\footnote{The symbols used in this section such as $\hat{I}$, $I$, $\hat{M}$, $M$ follow the definitions in section \ref{sec:rpn}.}:
\begin{equation}
\delta_{x}^{(t)} =
\begin{cases}
2(u_x^{(t)} - l_{d,s,t}(v)) & \mbox{if $x$ is a output node calculating $b_{d,s,t}(v)$} \\
0 & \mbox{otherwise}
\end{cases}
\end{equation}

Suppose node  $x$ is an activation node and $f(\cdot) = softclip(\cdot)$, let $y=\jmath_x$,
\begin{equation}
\begin{aligned}
\delta_{y}^{(t)}&=\frac{\partial {\cal L}}{\partial u_y^{(t)}}\\
&=\frac{\partial {\cal L}}{\partial u_x^{(t)}} \frac{\partial u_x^{(t)}}{\partial u_y^{(t)}}\\
&=\delta_{x}^{(t)} \frac{\partial f(u_y^{(t)})}{\partial u_y^{(t)}}
\end{aligned}
\end{equation}

Suppose node  $x=(d, i)$ is a sum node, then when node  $x$ passes its error, the error of node  $y \in \hat{I}_x$ is updated as
\begin{equation}
\begin{aligned}
\delta_{y}^{(t)} &=\delta_{y}^{(t)} + \frac{\partial {\cal L}}{\partial u_{x}^{(t)}} \frac{\partial u_{x}^{(t)}}{\partial u_{y}^{(t)}}\\
&=\delta_{y}^{(t)} +\delta_{x}^{(t)} \hat{w}_{x, y}
\end{aligned}
\end{equation}

Similarly, error of node  $y \in I_{x}$ is updated as
\begin{equation}
\begin{aligned}
\delta_{y}^{(t)} &= \delta_{y}^{(t)} + \frac{\partial {\cal L}}{\partial u_{x}^{(t)}} \frac{\partial u_{x}^{(t)}}{\partial u_{y}^{(t-1)}}\\
&=\delta_{y}^{(t)} +\delta_{x}^{(t)}{w}_{x, y}
\end{aligned}
\end{equation}

Suppose node  $x=(d, i)$ is a product node, then when node  $x$ passes its error, error of node  $y \in \hat{I}_x$ is updated as
\begin{equation}
\begin{aligned}
\delta_{y}^{(t)} &= \delta_{y}^{(t)} + \frac{\partial {\cal L}}{\partial u_{x}^{(t)}} \frac{\partial u_{x}^{(t)}}{\partial u_{y}^{(t)}}\\
&=\delta_{y}^{(t)} +\delta_{x}^{(t)} \hat{M}_{x, y} {u_y^{(t)}}^{\hat{M}_{x, y} - 1}\prod_{z \in \hat{I}_x - \{y\}} {u_z^{(t)}}^{\hat{M}_{x, z}} \prod_{z \in I_{x}} {u_z^{(t-1)}}^{{M}_{x, z}}
\end{aligned}
\end{equation}

Similarly, error of node  $y \in I_{x}$ is updated as
\begin{equation}
\begin{aligned}
\delta_{y}^{(t)} &= \delta_{y}^{(t)} + \frac{\partial {\cal L}}{\partial u_{x}^{(t)}} \frac{\partial u_{x}^{(t)}}{\partial u_{y}^{(t-1)}}\\
&=\delta_{y}^{(t)} +\delta_{x}^{(t)} M_{x, y} {u_y^{(t-1)}}^{M_{x, y} - 1}\prod_{z \in \hat{I}_{x} } {u_z^{(t)}}^{\hat{M}_{x, z}} \prod_{z \in {I}_{x} - \{y\}} {u_z^{(t-1)}}^{{M}_{x, z}}
\end{aligned}
\end{equation}

\end{document}